\documentclass[11pt,a4paper]{article}
\usepackage[hyperref]{emnlp2020}
\usepackage{times}
\usepackage{booktabs}
\usepackage{latexsym}
\usepackage{changes}

\usepackage{microtype}

\usepackage{tikz}
\usepackage{graphicx}
\usepackage{tikz-dependency}

\aclfinalcopy

\title{Creating a Universal Dependencies Treebank of Spoken Frisian-Dutch Code-switched Data}

\author{Anouck Braggaar \\
  University of Groningen \\
  \texttt{a.r.y.braggaar@student.rug.nl} \\\And
  Rob van der Goot \\
  IT University of Copenhagen \\
  \texttt{robv@itu.dk} \\}

\date{}

\begin{document}
\maketitle
\begin{abstract}
This paper explores the difficulties of annotating transcribed spoken
Dutch-Frisian code-switch utterances into Universal Dependencies. We make use
of data from the FAME! corpus, which consists of transcriptions and audio data.
Besides the usual annotation difficulties, this dataset is extra challenging
because of Frisian being low-resource, the informal nature of the data,
code-switching and non-standard sentence segmentation.  As a starting point,
two annotators annotated 150 random utterances in three stages of 50
utterances. After each stage, disagreements where discussed and resolved. An
increase of 7.8 UAS and 10.5 LAS points was achieved between the first and
third round. This paper will focus on the issues that arise when annotating a
transcribed speech corpus. To resolve these issues several solutions are
proposed.
\end{abstract}

\section{Introduction}
A key-component to developing low-resource dependency parsers for a specific
language-type, is an evaluation treebank.  In this paper we will focus on the
low-resource language West Frisian which is spoken in the Netherlands. We have
used the FAME!-project dataset which was created out of broadcasts from Omrop
Fryslân (Frisian radio broadcaster) \citep{yilmaz2016longitudinal}. Not only
are we dealing here with a low-resource language, we are also dealing with a
spontaneous speech dataset that contains code-switching between Frisian and
Dutch (the main language spoken in the Netherlands). As
\citet{yilmaz2016longitudinal} also mention, code-switching often occurs due to
the influence of Dutch.

In this paper we will elaborate on the issues that arise with annotating
Universal Dependencies~\cite{nivre-etal-2020-universal} on spoken code-switched
language. As a starting point, we have randomly selected 150 utterances that
contain at least one code-switching point from the FAME! corpus. We have
followed the utterance segmentation as can be found in the corpus.  The
code-switches are also already annotated in these utterances by the annotators
of the corpus \citep{yilmaz2016longitudinal}.  The final goal is to use the
annotated data to evaluate a dependency parser for this low-resource (and in
this case spoken) language. In this paper we will discuss issues that arose
during the annotation and propose possible solutions for these issues.

\section{Related work}
We are not the first to annotate spoken data. Previous work has annotated
English for conversation agents \citep{davidson-etal-2019-dependency},
Slovenian data \citep{dobrovoljc-martinc-2018-er}, Komi-Zyrian
\citep{partanen-etal-2018-first} and Turkish-German
\citep{cetinoglu-coltekin-2019-challenges}. Commonly mentioned problems are
disfluencies and sentence segmentation \citep{dobrovoljc-martinc-2018-er}. Two
main types of solutions can be identified; adapting the existing guidelines
\citep{cetinoglu-coltekin-2019-challenges} versus extending them
\citep{davidson-etal-2019-dependency}.

Previous research also focuses on creating treebanks for code-switch data.
\citet{cetinoglu-coltekin-2019-challenges} focus on the issues that arise when
annotating a spoken Turkish-German code-switch treebank and make a distinction
between issues that are code-switch specific or related to spoken language.
They conclude that they use dependencies that rarely occur in monolingual
Turkish or German treebanks. \citet{seddah-etal-2020-building} create a
treebank for an Arabic dialect which contains a high amount of code-switching
and language variation.  \citet{partanen-etal-2018-first} create a spoken
treebank for Komi-Zyrian with code-switching to Russian. They argue that some
language-specific issues might be difficult to fully address with Universal
Dependencies.

\section{Annotations}
Overall, we tried to closely follow the existing Universal Dependency
guidelines and the existing annotations of the Dutch Alpino and LassySmall
treebanks~\cite{van2002alpino, van2013large}. But as we will show, some
phenomena in spoken language may not be easy to annotate with an appropriate
label.

\subsection{Inter Annotator Agreements}
We have both annotated in total 150 randomly selected utterances from scratch.
After every batch of 50 sentences we discussed issues and adjusted our
annotation scheme.  We report accuracy over Universal Parts-of-speech tags,
Unlabelled Attachment Score (UAS) and Labelled Attachment Score
(LAS)~\cite{zeman-etal-2018-conll} in Table~\ref{tab:agreement}. Even though
the scores improve over time, the final agreements are still relatively low;
previous work on social media data reached an LAS of
84~\cite{liu-etal-2018-parsing}, and for code-switched data an LAS of 92 is
reported~\cite{bhat-etal-2017-joining}. 

We found that the four main sources of disagreement where due to 1)
difficulties in ungrammatical constructions 2) sentence segmentation 3)
interpretation of the utterances (ambiguity) 4) annotators had to learn the
guidelines. In fact, there were very few issues with the code-switch aspect of
this data.  The reason for this could be that only very short parts of the
utterance (e.g. only a content word) are switched and that there is a high
degree of resemblance between the two languages \citep{wolf1996structural},
making the switches not directly an issue while annotating.  In the following
section, we will discuss how we overcame issues in the first and second sources
of disagreement.

\begin{table}
\centering
\begin{tabular}{l r r r }
\toprule
  & POS & UAS & LAS \\
\midrule
 Round 1 & 69.5 & 72.3 & 60.9 \\
 Round 2 & 87.1 & 76.1 & 64.6 \\
 Round 3 & 89.7 & 80.1 & 71.4 \\ 
\bottomrule
\end{tabular}
\caption{POS, UAS and LAS scores between the two annotators.}
\label{tab:agreement}
\end{table}

\subsection{Annotation Issues}
In this section we will discuss the two most common sources of disagreement we
encountered.

This first example shows a phenomenon that occurred often and was often
annotated differently:
\begin{figure}[htp]
    \centering
    \resizebox{\linewidth}{!}{
 \begin{dependency}
\begin{deptext}
benammen	\& eh	\& foarsitter	\& eh	\& Van	\& Raaij	\& dy	\& hat	\& eh	\& oare	\& plannen	\\
especially  \& eh \& chairman \& eh \& Van \& Raaij \& whom \& has \& eh \& other \& plans \\
\end{deptext}
\depedge{3}{1}{amod}
\depedge{3}{2}{discourse}
\depedge[edge unit distance=2ex]{8}{3}{nsubj}
\depedge{3}{4}{discourse}
\depedge{3}{5}{appos}
\depedge{5}{6}{flat:name}
\depedge{8}{7}{expl}
\deproot[edge unit distance=4ex]{8}{root}
\depedge{8}{9}{discourse}
\depedge{11}{10}{amod}
\depedge{8}{11}{obj}
\end{dependency}}
\end{figure}

A common source of confusion was when something/someone is referred to by name
(in this case "foarsitter Van Raaij") and it is later referred to again with a
relative pronoun ("dy"). There are multiple ways to annotate this. Our first
choice was to annotate "foarsitter Van Raaij" as being the subject and "dy" as
being in a determiner relation. A second option would be to annotate "dy" as
the subject of the sentence and the other part as being in an appos relation,
defining the "dy". Eventually, we chose to annotate "dy" as expletive and keep
"foarsitter Van Raaij" as the subject.

The second example shows a couple of different issues that were specific to
spoken language:

\begin{figure}[htp]
    \centering
    \resizebox{\linewidth}{!}{
\begin{dependency}
   \begin{deptext}
     hoe \& dan \& ek \& jongens \& dy \& moties \& dy \& eh \& dy \& moatte \& der \& trochkomme \& en \\
     \& anyways \& \& guys \& those \& motions \& they \& eh \& they \& have \& there \& come through \& and \\
   \end{deptext}
\depedge[edge unit distance=1.85ex]{12}{1}{discourse}
\depedge{1}{2}{fixed}
\depedge{1}{3}{fixed}
\depedge[edge unit distance=2ex]{12}{4}{vocative}
\depedge{6}{5}{det}
\depedge[edge unit distance=2ex]{12}{6}{nsubj}
\depedge{9}{7}{reparandum}
\depedge{9}{8}{discourse}
\depedge{12}{9}{expl}
\depedge{12}{10}{aux}
\depedge{12}{11}{advmod}
\deproot[edge unit distance=5.5ex]{12}{root}
\depedge{12}{13}{dislocated}
\end{dependency}
}
\end{figure}

The first thing to notice is that it starts with "hoe dan ek" which is an
expression that is mainly used in spoken speech. Therefore we decided to label
this as discourse.  The most striking phenomenon in this utterance is the fact
that it doesn't seem to have a proper ending, it seems to go on "en" ("and").
This is something that happens a lot because of the spoken data and because of
segmentation. We have chosen to attach these elements to the root with the
dislocated label. Normally you would have elements on the right to which most
of these dislocated elements would attach.

\section{Conclusion}
In this paper we have discussed some of the issues that arise when annotating a
code-switch spoken treebank. As we have discussed we follow the general
Universal Dependency guidelines and existing Dutch annotations. Annotating is
still work in progress and our LAS and UAS scores leave room for improvement.
Eventually we would like to annotate a larger amount of utterances that can be
used to evaluate a dependency parser in a low-resource setup.

\bibliographystyle{acl_natbib}
\bibliography{emnlp2020}

\end{document}